\documentclass{article}
\usepackage{spconf,amsmath,graphicx}

\usepackage{epsfig} 
\usepackage{fancyhdr} 
\usepackage{longtable} 
\usepackage{color} 
\usepackage{amsmath} 
\usepackage{amssymb} 
\usepackage{bm} 
\usepackage{psfrag} 
\usepackage{url} 
\usepackage[linktocpage]{hyperref} 
\usepackage{graphicx} 
\usepackage{tabularx}
\usepackage[table,xcdraw]{xcolor}

\usepackage{algorithm}
\usepackage{algpseudocode}
\usepackage{chngcntr}
\usepackage{multirow}
\usepackage{comment}

\usepackage{mathtools}
\usepackage{subcaption}

\usepackage{lipsum}  

\definecolor{ashgrey}{rgb}{0.7, 0.75, 0.71}

\newcommand{\fref}[1]{Fig.~\ref{#1}}
\newcommand{\sref}[1]{Section~\ref{#1}}
\newcommand{\cref}[1]{Chapter~\ref{#1}}

\newcommand{\tref}[1]{Table~\ref{#1}}
\newcommand{\eref}[1]{Eq.~\ref{#1}}

\title{LT-ViT: A Vision Transformer for multi-label Chest X-ray classification}

\name{Umar Marikkar$^{1}$ \qquad Sara Atito$^{1,2}$ \qquad Muhammad Awais$^{1,2}$ \qquad Adam Mahdi$^{3}$}
\address{$^{1}$Surrey Institute of People Centered AI,
	University of Surrey,\\
        $^{2}$Centre of Vision, Speech and Signal Processing (CVSSP),
	University of Surrey, \\
        $^{3}$Oxford Internet Institute, University of Oxford}
 
\begin{document}
%
\maketitle
\begin{abstract}

Vision Transformers (ViTs) are widely adopted in medical imaging tasks, and some existing efforts have been directed towards vision-language training for Chest X-rays (CXRs). However, we envision that there still exists a potential for improvement in vision-only training for CXRs using ViTs, by aggregating information from multiple scales, which has been proven beneficial for non-transformer networks.  Hence, we have developed LT-ViT, a transformer that utilizes combined attention between image tokens and randomly initialized auxiliary tokens that represent labels. Our experiments demonstrate that LT-ViT (1) surpasses the state-of-the-art performance using pure ViTs on two publicly available CXR datasets, (2) is generalizable to other pre-training methods and therefore is agnostic to model initialization, and (3) enables model interpretability without grad-cam and its variants.

\end{abstract}
\section{Introduction}
\label{sec:intro}

With the advent of Vision Transformers (ViTs) \cite{dosovitskiy2020image} for computer vision tasks, many studies have applied ViTs to medical images, including 2D images—radiographs \cite{pang2022popar} and 3D volumes—MRI scans \cite{hatamizadeh2022swin}. More recently, Chest X-Ray (CXR) based image classification benchmarks have been fuelled by vision-language training, where CXRs use their corresponding radiology reports as labels \cite{wangmulti, wang2022medclip, muller2022joint}.

The attention mechanism inherently present in the transformer architecture \cite{vaswani2017attention} allows for learnable interaction between tokens (feature vectors) in such a way that a single token can process information from any other token within or out of the same domain. Leveraging this, BERT \cite{devlin2018bert} introduced a class token $\mathbf{[CLS]}$, which aims to aggregate information from, and propagate information to all the domain-specific data tokens. Subsequently, the common practice in the vision community is to utilize this class token that aggregates the global information to predict the output of the given task, e.g., multi-class classification \cite{dosovitskiy2020image}.


Unlike multi-class classification problems, it is imperative for the multi-label classification tasks that the network is able to understand the connections between each label, such that it is able to infer information about a specific label from another. Using the transformer architecture, C-Tran \cite{lanchantin2021general} proposes a transformer encoder to model dependencies between labels and data, whilst Query2label \cite{liu2021query2label} utilizes a transformer decoder to do so. Both these studies use auxiliary tokens \cite{sandler2022fine,xu2022groupvit,carion2020end} to model dependencies,  however, C-Tran enables bi-directional attention between data and labels, and Query2label is built upon a framework where the auxiliary tokens only attend to fully encoded data tokens. For the former, this results in data tokens having to be transferred into a data-image hybrid space, thereby damaging its domain-specificity, and in the latter, the auxiliary tokens only attend to one set of data tokens, thereby preventing them from learning from the data at multiple scales. 

A shortcoming noted from the aforementioned studies is that they use a full vision backbone and then apply the auxiliary tokens to model dependencies. However, we note that for medical images, information aggregation from multiple scales is predominant for better multi-label classification performance \cite{WANG2021101846}. To this end, we propose \textbf{L}abel \textbf{T}oken \textbf{Vi}sion \textbf{T}ransformer—\textbf{LT-ViT}, a simple transformer architecture designed for multi-label CXR classification. LT-ViT enables joint learning of auxiliary label and data tokens, where the label tokens attend to data tokens within the actual vision backbone. This enables multi-scale learning and avoids extra layers of compute. Using LT-ViT, we propose to improve vision-only training using transformers for CXRs, that would result in the potential generalizability towards non-CXR medical images, where image-text paired data is not present. This paper aims to show the following: 

\begin{enumerate}
    \item LT-ViT surpasses state-of-the-art performance using pure ViTs on two publicly available CXR datasets (\sref{sec:GMML_results}). 
    \item LT-ViT is generalizable to other pre-training methods and therefore is agnostic to model initialization (\sref{sec:LT_general}).
    \item LT-ViT enables model interpretability without grad-cam \cite{selvaraju2017grad} and its variants (\sref{sec:analysis}).
\end{enumerate}


\begin{figure*}[h!]
    \centering
    \includegraphics[width=0.70\textwidth]{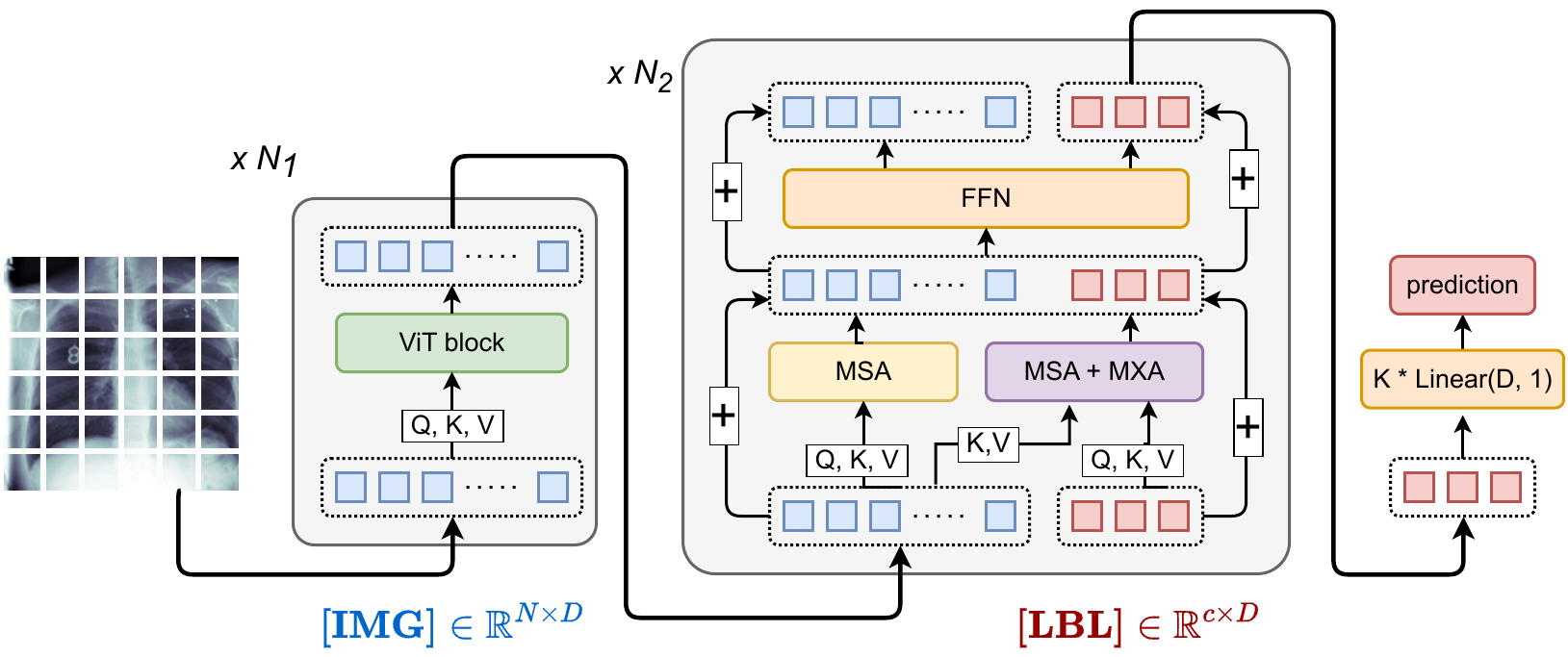}
    \caption{LT-ViT architecture using auxiliary label tokens. [+]—skip connection. MSA—multi-head self-attention, MXA—multi-head cross-attention. QKV denote Query, Key, and Value. In the LT-ViT blocks, $\mathbf{[LBL]}$ act as queries to all tokens within the network, whilst the $\mathbf{[IMG]}$ act as queries only to $\mathbf{[IMG]}$.}
    \label{fig:label_tokens}
\end{figure*}
\pagebreak

\section{Methodology}
\label{sec:method}

In this paper, we propose LT-ViT, a generic framework for multi-label medical image classification that leverages ViTs to exploit the dependencies among visual features and labels, which therefore improve the explainability of the predictions. We briefly explain vision transformers in Section \ref{sec:ViT}. Next, in Section \ref{sec:lt_vit}, we explain the overall architecture of LT-ViT for multi-label classification.

\subsection{Vision Transformers (ViTs)}
\label{sec:ViT}
ViT \cite{dosovitskiy2020image} receives as input a sequence of patches obtained by tokenizing the input image $\mathbf{x} \in \mathbb{R}^{H \times W \times C}$ into $n$ flattened $2D$ patches of size $p \times p \times C$ pixels, where $H$, $W$, and $C$ are the height, width, and the number of channels of the input image and $n$ is the total number of patches. Each patch is then projected with a linear layer to $D$ hidden dimensions. In order to retain the relative spatial relation between the patches, learnable position embeddings are added to the patch embeddings as an input to the transformer encoder. Further, a class token $\mathbf{[CLS]} \in \mathbb{R}^{1 \times D}$ is concatenated to the data tokens, where its output is used to represent the entire image. The standard transformer encoder consists of $L$ consecutive multi-head self-attention and multi-layer perceptron blocks.  

\subsection{LT-ViT framework}
\label{sec:lt_vit}

For general single/multi label classification tasks involving ViTs, the output of the class token $\mathbf{[CLS]}$ is fed to a prediction head, which projects it to $c$ nodes corresponding to the number of classes in the given dataset. However, in a multi-label setting, $\mathbf{[CLS]}$ contains the global representation of all the true labels in the image. Therefore, the representations for each label are not directly separable. To alleviate this problem, we introduce learnable label tokens (as  $Y \in \mathbb{R}^{c \times D}$ to the network, where $c$ is the total number of labels in the dataset. The proposed label tokens possess the ability to update themselves by selectively attending to both image and label tokens.

Specifically, at layer $l$, for each image token set $X^{l} \in \mathbb{R}^{n \times D}$, and label token set $Y^{l} \in \mathbb{R}^{c \times D}$, we project these tokens to their corresponding query, key and value vectors by,

\begin{equation}
    \begin{bmatrix}
    X_{Q} & X_{K} & X_{V}\\
    Y_{Q} & Y_{K} & Y_{V}
    \end{bmatrix} = \begin{bmatrix}
    X \\ Y
    \end{bmatrix}\cdot W_{QKV}\quad,
\end{equation}
where $W_{QKV} \in \mathbb{R}^{D \times 3D}$ is the learnable projection matrix. Then the image-label dependencies are computed, and their tokens are updated as,

\begin{equation}
\label{eq:cross_att}
    \begin{bmatrix}
    \Tilde{X} \\ \Tilde{Y}
    \end{bmatrix}
    = FFN \left(
    \begin{bmatrix}
    attn\left(X_{Q}, X_{K}, X_{V}\right) \\
    attn\left(Y_{Q},
    \begin{bmatrix}
    X_{K} \\ Y_{K}
    \end{bmatrix}, \begin{bmatrix}
    X_{V} \\ Y_{V}
    \end{bmatrix}\right)
    \end{bmatrix}
    \right) \quad,
\end{equation}
where $\Tilde{X}$ and $\Tilde{Y}$ are the updated image and label tokens, while $FFN(\cdot)$ and $attn(Q,K,V)$ denote the feed-forward and multi-head attention computations as in \cite{vaswani2017attention}. \eref{eq:cross_att} ensures the one-way flow of information from image to label tokens, allowing each label token to learn from the data, while also allowing the sharing of information between label tokens through self-attention, as needed. From here on, we will refer to the label tokens $Y$ as $\mathbf{[LBL]}$.

The schematic of the LT-ViT architecture is shown in Figure \ref{fig:label_tokens}. The label tokens can be introduced at any point in the ViT, hence $N_{1}$ and $N_{2}$ are adjustable, under the constraint $N_{1} + N_{2} = L$, where $L$ is the depth of the original ViT. Therefore, apart from the newly initialized label tokens, all learnable parameters can be directly transferred from any pre-trained model. 

After $N_2$ LT-ViT blocks, each updated label token $\mathbf{[LBL]}^{k}$, where $k \coloneqq \{1, \dots, c\}$, is passed through a linear layer with output dimensionality equal to one, and so the process is formulated as $c$ binary classification problems, following \cite{liu2021query2label}. The binary cross-entropy loss (BCELoss) for each label is then computed between the predicted value and the ground truth logit in order to back-propagate through the network.

\section{Experiments and Implementation}
\label{sec:experiments}

LT-ViT is a generic framework for multi-label classification that can be trained from scratch (i.e., random initialization), or initialized from a pre-trained model from in-domain or out of domain data. To show the effectiveness of our proposed framework, we followed both scenarios and compare the performance with the state-of-the-art. We provide information about the employed datasets and experiments, and the implementation details in \sref{sec:dataset} and \sref{sec:imp_details}, respectively. In \sref{sec:results}, we discuss the performance and analysis of the proposed method.

\subsection{Datasets and Experiments}
\label{sec:dataset}
For pre-training, we used the NIH-CXR14 dataset \cite{wang2017chestx} which contains 112,120 frontal-view Chest X-ray images from 30,805 patients. We used the official 80/20 train/test split for all experiments. For fine-tuning and evaluation, we employed two datasets: NIH-CXR14 and CheXpert \cite{irvin2019chexpert}. The latter is an additional large scale CXR dataset consisting of 224,316 X-ray images from 65,240 patients. It contains an expertly annotated test set of 234 radiographs, which we have used as our default test set. We evaluate our methods by performing experiments on both the 5-label (the most commonly used) and 13-label versions of this dataset, to examine the effects of LT on the number of labels used.    

We employed Group Masked Model Learning (GMML) \cite{GMML} as our pre-training method, as it exhibits strong performance on small datasets, notably on radiographs \cite{anwar2022ss, atito2022sb}. A simple ViT-S pre-trained using GMML was set as our baseline, and this baseline was compared with existing state-of-the-art pre-training methods \cite{haghighi2022dira, pang2022popar} on CXR's to outline the performance delta needed to be fulfilled by LT-ViT. Following that, we compared the performance change obtained upon introducing LT-ViT versus existing benchmarks that use a larger number of trainable parameters. 

Furthermore, we investigated the generalizability of our proposed label tokens by obtaining models pre-trained on different domains, specifically, MIMIC-CXR \cite{johnson2019mimic} (Vision and Language CXR) and ImageNet \cite{ILSVRC15} (Natural images). We pre-trained a ViT-S on MIMIC-CXR using MGCA \cite{wangmulti}, and obtained a pre-trained checkpoint on ImageNet from the official DINO \cite{caron2021emerging} repository. We introduced LT-ViT on these pre-trained checkpoints and compared their performance on the fine-tuning datasets.

\subsection{Implementation details}
\label{sec:imp_details}
To obtain the pre-trained ViT on NIH-CXR8, we pre-trained a ViT-S using GMML on 4 RTX 3090 GPU's with a per-gpu batch size of 64, AdamW optimizer with a learning rate of $5e^{-4}$ and a weight decay of 0.04.

In the implementation of LT-ViT,  the $N_2$ value (number of layers carrying label tokens) was set to 4, based on a search within the range [2:2:12]. We expanded on the official DeiT \cite{touvron2021training} code base for the fine-tuning experiments and ran them on a single RTX 3090 GPU with batch size 64. We used the Adam optimizer with a learning rate of $1e^{-5}$ for CheXpert and $2.5e^{-6}$ for NIH-CXR14. Q2L \cite{liu2021query2label}, was trained using the implementation details provided in their study, along with a quick hyper-parameter search for the downstream datasets.

\section{Results}
\label{sec:results}

The potential of LT-ViT to reach state-of-the-art CXR classification benchmarks using transformers is outlined in \sref{sec:GMML_results}. Following that, we put forward the generalizability of LT-ViT and the qualitative analysis of our method in \sref{sec:LT_general} and \sref{sec:analysis}, respectively.

\subsection{LT-ViT obtains state-of-the-art performance on CXR classification with ViTs using a fraction of parameters}
\label{sec:GMML_results}

\tref{tab:pretrain} presents the comparison between the proposed method and previously established multi-label classification methods using transformers. 

\begin{table}[htb]
\caption{Comparison of LT-ViT vs. existing benchmarks pre-trained on NIH-CXR14. }
\label{tab:pretrain}

\resizebox{\columnwidth}{!}{%
\begin{tabular}{l|ccc|ccc}
\hline
\multicolumn{1}{c|}{\multirow{2}{*}{Method}} & \multicolumn{1}{c}{\multirow{2}{*}{Model}} & \multirow{2}{*}{Params} & \multirow{2}{*}{Img.res} & \multicolumn{3}{c}{AUC (\%)}                                                                             \\ \cline{5-7} 
\multicolumn{1}{c|}{}                                 & \multicolumn{1}{c}{}                                &                                                                                &                                    & \multicolumn{1}{c}{NIH-14} & \multicolumn{1}{c}{CheX-05} & \multicolumn{1}{c}{CheX-13} \\ \hline
DIRA \cite{haghighi2022dira}            & Resnet50      & 23M       & $224^2$      & \multicolumn{1}{c}{81.12}            & \multicolumn{1}{c}{87.59}             & -                                    \\ \hline
POPAR-3 \cite{pang2022popar}            & ViT-B         & 86M       & $224^2$      & \multicolumn{1}{c}{79.58}            & \multicolumn{1}{c}{87.86}             & -                                    \\
POPAR                                   & Swin-B        & 88M       & $448^2$      & \multicolumn{1}{c}{81.81}            & \multicolumn{1}{c}{88.34}             & -                                    \\ \hline
GMML                                    & ViT-S         & 21M       & $224^2$      & \multicolumn{1}{c}{81.28}            & \multicolumn{1}{c}{87.36}             & 73.65                                \\
+ Q2L \cite{liu2021query2label}    & ViT-S         & 26M       & $224^2$      & \multicolumn{1}{c}{81.09}            & \multicolumn{1}{c}{88.12}             & 77.12                                \\
\textbf{+ LT-ViT (ours)}                       & ViT-S         & 21M       & $224^2$      & \multicolumn{1}{c}{\textbf{81.98}}   & \multicolumn{1}{c}{\textbf{88.90}}    & \textbf{77.34}                       \\ \hline
\end{tabular}
}
\end{table}

From \tref{tab:pretrain}, we find that GMML, in isolation, exhibit comparable performance relative to DIRA and POPAR-3 under the same image resolution and therefore is suitable for our ViT baseline. Nonetheless, we note that as the image input resolution increases, the resulting increase in fine-grained structural detail leads to improved accuracy, as observed for POPAR.

Our proposed LT-ViT, when applied to the ViT baseline, is able to outperform POPAR, while maintaining the original input resolution and requiring only a negligible increase in the number of trainable parameters, caused by the token embedding for each label. We also find that LT-ViT outperforms the existing multi-label training paradigm Query2Label (Q2L) on Transformers.

\subsection{LT-ViT is agnostic to model initialization}
\label{sec:LT_general}

\tref{tab:LT_generalizable} shows the results obtained by using LT under various initialization states/pre-trained datasets. In addition to pre-training using NIH-CXR14 as in \sref{sec:GMML_results}, we evaluate our method on random initialization, and on models pre-trained on two distinct datasets. 

\begin{table}[]
\caption{Study of LT-ViT generalizability over model initializations. \textbf{rnd} indicates random initialization (no pre-training).}
\label{tab:LT_generalizable}

\resizebox{\columnwidth}{!}{%
\begin{tabular}{lc|ccc}
\cline{2-5}
\multirow{2}{*}{}      & \multirow{2}{*}{\begin{tabular}[c]{@{}c@{}}Pre-train\\ dataset\end{tabular}} & \multicolumn{3}{c}{AUC (\%)}\\ \cline{3-5} 
&& NIH-CXR14 &  CheXpert-5 & CheXpert-13 \\ \hline
\multicolumn{1}{l|}{rnd} & \multirow{2}{*}{-} & 72.37 & 83.30 & 68.05 \\
\multicolumn{1}{l|}{\textbf{+ LT-ViT}}  & &                  \textbf{73.32} & \textbf{85.90} & \textbf{70.19} \\ \hline
\multicolumn{1}{l|}{DINO \cite{caron2021emerging}}           & \multirow{2}{*}{ImageNet-1k} & 80.79 &  87.38 & 73.23\\
\multicolumn{1}{l|}{\textbf{+ LT-ViT}} & &
\textbf{81.11} & \textbf{88.44}& \textbf{75.43}\\ \hline
\multicolumn{1}{l|}{MGCA \cite{wangmulti}} & 
\multirow{2}{*}{MIMIC-CXR}                                   & 81.45 & 87.98 & 71.88  \\
\multicolumn{1}{l|}{\textbf{+ LT-ViT}} && 
\textbf{81.87} & \textbf{88.32} & \textbf{72.13} \\ \hline
\end{tabular}
}
\end{table}

Based on \tref{tab:LT_generalizable}, we have found that the proposed LT method improves the performance of multi-label classification across all initialization states and fine-tuning datasets. We observe that a weaker initialization (random) leads to a higher performance increase compared to stronger initializations from ImageNet \cite{ILSVRC15} and MIMIC-CXR \cite{johnson2019mimic}..

These findings provides evidence that LT-ViT can be applied to any ViT model for multi-label CXR classification. These results could suggest the possibility of extending LT-ViT to other multi-label tasks using transformers.

\subsection{Analysis of our framework}
\label{sec:analysis}

\subsubsection{Visualization}

\fref{fig:label_tokens} illustrates the smoothed attention maps of a specific label token in relation to the visual feature tokens, as well as the bounding box annotation for the corresponding pathology. We observe from the visualization that the label token is capable of accurately localizing the affected area for the given pathology, as shown by the ground truth annotation, represented by the red box. In contrast, when using the commonly used $\mathbf{[CLS]}$ token (as performed in \cite{caron2021emerging}), the attention map provides a union of all pathologies present in the X-ray, lacking the specificity achieved by the label token.

\begin{figure}[]
    \centering
    \includegraphics[width=\columnwidth]{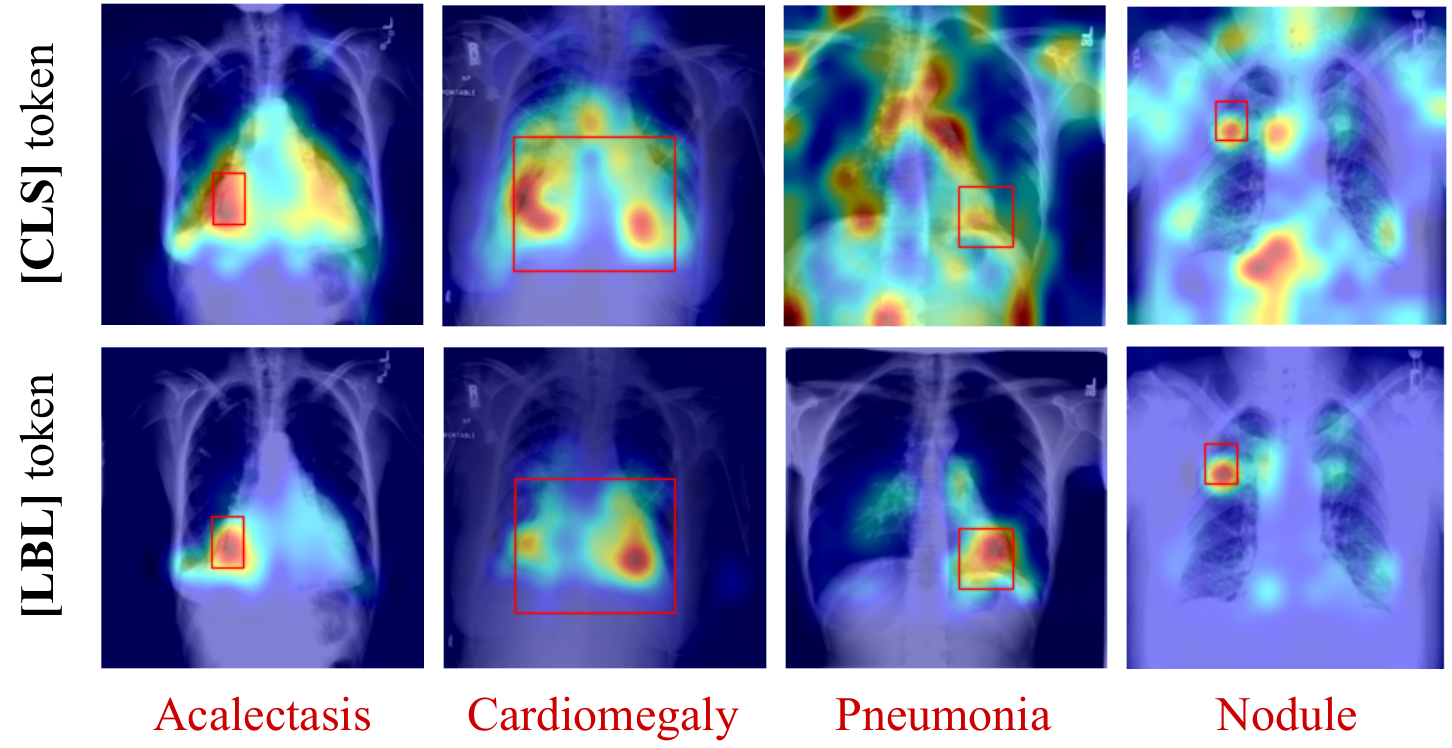}
    \caption{Visualization of attention maps for $\mathbf{[CLS]}$ and $\mathbf{[LBL]}$ tokens. Red text indicates the pathology denoted by the bounding box ground truth.}
    \label{fig:label_tokens}
\end{figure}

\subsubsection{Ablations}

We conducted an ablation study to investigate the impact of the label token related attention components within the network architecture, as seen in \tref{tab:attention}. Firstly, we employed full self-attention, which is equivalent to having '$\mathbf{c}$' amounts of $\mathbf{[CLS]}$ tokens and is architecturally similar to C-Tran \cite{lanchantin2021general}. Secondly, we removed the self-attention component within the label token space, which means that each label token only attends to the image tokens. This implies that no information is shared between label tokens during the forward pass. 

\begin{table}[htb]
\caption{CheXpert-5 performance on different interactions between image and label tokens. [IMG], [LBL] indicate image and label tokens respectively.}
\label{tab:attention}
\resizebox{\columnwidth}{!}{%
\begin{tabular}{cc|cc|c}
\hline
\multicolumn{2}{c|}{Image tokens} & \multicolumn{2}{c|}{Label tokens} & \multirow{2}{*}{AUC (\%)} \\ \cline{1-4}
Query           & Key, Val       & Query           & Key, Val       &                                    \\ \hline
{[}LBL{]}{[}IMG{]}   & {[}LBL{]}{[}IMG{]}  & {[}LBL{]}{[}IMG{]}   & {[}LBL{]}{[}IMG{]}  & 88.47                              \\
{\textbf{[IMG]}}            & {\textbf{[IMG]}}           & {\textbf{[LBL]}}            & {\textbf{[LBL][IMG]}}  & \textbf{88.90}                     \\
{[}IMG{]}            & {[}IMG{]}           & {[}LBL{]}            & {[}IMG{]}           & 88.44                              \\
{[}IMG{]}            & {[}IMG{]}           & n/a                  & n/a                 & 87.36                              \\ \hline
\end{tabular}%
}
\end{table}

We note the significance of one-way attention in impeding data tokens learning from label tokens in \tref{tab:attention}, as shown by the slight decrease in performance when utilizing full self-attention. Furthermore, it is imperative that label tokens are permitted to attend to other label tokens to capture the inter-label dependencies, as indicated by the decrease in accuracy when there is no self-attention between label tokens.

\section{Conclusion}
\label{sec:conclusion}

We proposed \textbf{LT-ViT}, a simple transformer-based framework for multi-label classification in CXRs. We showed that, without any bells and whistles during training, LT-ViT is able to increase multi-label classification performance consistently throughout the tested datasets. We also discovered that LT-ViT outperforms the existing multi-label training paradigm when applied to CXRs, without the use of additional decoder layers. Furthermore, we outlined the generalizability over numerous model initializations, and found out that the weaker the initialization is, the more the model becomes dependent on LT-ViT to improve classification performance. In addition, we showed that the label tokens in LT-ViT are able to accurately localize the pathologies present in the CXR image. Given these promising results on CXR data, the next step will be to study the generalizability of LT-ViT across more downstream datasets, in both medical and non-medical domains. \textbf{Acknowledgements}
This work was supported by the EPSRC grants MVSE (EP/V002856/1) and JADE2 (EP/T022205/1).

\vfill\pagebreak

\bibliographystyle{IEEEbib}
\bibliography{refs}

\begin{thebibliography}{10}

\bibitem{dosovitskiy2020image}
Alexey Dosovitskiy, Lucas Beyer, Alexander Kolesnikov, et~al.,
\newblock ``An image is worth 16x16 words: Transformers for image recognition
  at scale,''
\newblock {\em ICLR}, 2021.

\bibitem{pang2022popar}
Jiaxuan Pang, Fatemeh Haghighi, DongAo Ma, et~al.,
\newblock ``Popar: Patch order prediction and appearance recovery for
  self-supervised medical image analysis,''
\newblock {\em DART, held in Conjunction with MICCAI}, 2022.

\bibitem{hatamizadeh2022swin}
Ali Hatamizadeh, Vishwesh Nath, Yucheng Tang, et~al.,
\newblock ``Swin unetr: Swin transformers for semantic segmentation of brain
  tumors in mri images,''
\newblock {\em BrainLes held in Conjunction with MICCAI}, 2022.

\bibitem{wangmulti}
Fuying Wang, Yuyin Zhou, Shujun Wang, Varut Vardhanabhuti, et~al.,
\newblock ``Multi-granularity cross-modal alignment for generalized medical
  visual representation learning,''
\newblock {\em NeurIPS}, 2022.

\bibitem{wang2022medclip}
Zifeng Wang, Zhenbang Wu, Dinesh Agarwal, and Jimeng Sun,
\newblock ``Medclip: Contrastive learning from unpaired medical images and
  text,''
\newblock {\em EMNLP}, 2022.

\bibitem{muller2022joint}
Philip M{\"u}ller, Georgios Kaissis, Congyu Zou, et~al.,
\newblock ``Joint learning of localized representations from medical images and
  reports,''
\newblock {\em ECCV}, 2022.

\bibitem{vaswani2017attention}
Ashish Vaswani, Noam Shazeer, Niki Parmar, et~al.,
\newblock ``Attention is all you need,''
\newblock {\em NeurIPS}, 2017.

\bibitem{devlin2018bert}
Jacob Devlin, Ming-Wei Chang, Kenton Lee, et~al.,
\newblock ``Bert: Pre-training of deep bidirectional transformers for language
  understanding,''
\newblock {\em NAACL}, 2018.

\bibitem{lanchantin2021general}
Jack Lanchantin, Tianlu Wang, Vicente Ordonez, et~al.,
\newblock ``General multi-label image classification with transformers,''
\newblock {\em CVPR}, 2021.

\bibitem{liu2021query2label}
Shilong Liu, Lei Zhang, Xiao Yang, et~al.,
\newblock ``Query2label: A simple transformer way to multi-label
  classification,''
\newblock {\em arXiv preprint arXiv:2107.10834}, 2021.

\bibitem{sandler2022fine}
Mark Sandler, Andrey Zhmoginov, Max Vladymyrov, et~al.,
\newblock ``Fine-tuning image transformers using learnable memory,''
\newblock {\em CVPR}, 2022.

\bibitem{xu2022groupvit}
Jiarui Xu, Shalini De~Mello, Sifei Liu, et~al.,
\newblock ``Groupvit: Semantic segmentation emerges from text supervision,''
\newblock {\em CVPR}, 2022.

\bibitem{carion2020end}
Nicolas Carion, Francisco Massa, Gabriel Synnaeve, et~al.,
\newblock ``End-to-end object detection with transformers,''
\newblock {\em ECCV}, 2020.

\bibitem{WANG2021101846}
Hongyu Wang, Shanshan Wang, Zibo Qin, et~al.,
\newblock ``Triple attention learning for classification of 14 thoracic
  diseases using chest radiography,''
\newblock {\em Medical Image Analysis}, 2021.

\bibitem{selvaraju2017grad}
Ramprasaath~R Selvaraju, Michael Cogswell, Abhishek Das, et~al.,
\newblock ``Grad-cam: Visual explanations from deep networks via gradient-based
  localization,''
\newblock {\em ICCV}, 2017.

\bibitem{wang2017chestx}
Xiaosong Wang, Yifan Peng, Le~Lu, et~al.,
\newblock ``Chestx-ray8: Hospital-scale chest x-ray database and benchmarks on
  weakly-supervised classification and localization of common thorax
  diseases,''
\newblock {\em CVPR}, 2017.

\bibitem{irvin2019chexpert}
Jeremy Irvin, Pranav Rajpurkar, Michael Ko, et~al.,
\newblock ``Chexpert: A large chest radiograph dataset with uncertainty labels
  and expert comparison,''
\newblock {\em AAAI}, 2019.

\bibitem{GMML}
Sara Atito, Muhammad Awais, and Josef Kittler,
\newblock ``Gmml is all you need,''
\newblock {\em arXiv preprint arXiv:2205.14986}, 2022.

\bibitem{anwar2022ss}
Syed~Muhammad Anwar, Abhijeet Parida, Sara Atito, et~al.,
\newblock ``Ss-cxr: Multitask representation learning using self supervised
  pre-training from chest x-rays,''
\newblock {\em arXiv preprint arXiv:2211.12944}, 2022.

\bibitem{atito2022sb}
Sara Atito, Syed~Muhammad Anwar, Muhammad Awais, et~al.,
\newblock ``Sb-ssl: Slice-based self-supervised transformers for knee
  abnormality classification from mri,''
\newblock {\em MILLanD held in Conjunction with MICCAI}, 2022.

\bibitem{haghighi2022dira}
Fatemeh Haghighi, Mohammad Reza~Hosseinzadeh Taher, Michael~B Gotway, et~al.,
\newblock ``Dira: discriminative, restorative, and adversarial learning for
  self-supervised medical image analysis,''
\newblock {\em CVPR}, 2022.

\bibitem{johnson2019mimic}
Alistair~EW Johnson, Tom~J Pollard, Nathaniel~R Greenbaum, et~al.,
\newblock ``Mimic-cxr-jpg, a large publicly available database of labeled chest
  radiographs,''
\newblock {\em arXiv preprint arXiv:1901.07042}, 2019.

\bibitem{ILSVRC15}
Olga Russakovsky, Jia Deng, Hao Su, Jonathan Krause, et~al.,
\newblock ``{ImageNet Large Scale Visual Recognition Challenge},''
\newblock {\em IJCV}, vol. 115, no. 3, pp. 211--252, 2015.

\bibitem{caron2021emerging}
Mathilde Caron, Hugo Touvron, Ishan Misra, Herv{\'e} J{\'e}gou, et~al.,
\newblock ``Emerging properties in self-supervised vision transformers,''
\newblock {\em ICCV}, 2021.

\bibitem{touvron2021training}
Hugo Touvron, Matthieu Cord, Matthijs Douze, et~al.,
\newblock ``Training data-efficient image transformers \& distillation through
  attention,''
\newblock {\em ICML}, 2021.

\end{thebibliography}

\end{document}